\documentclass{commtr}

\usepackage{algorithmic}
\usepackage{algorithm}
\usepackage{graphicx}
\usepackage{amsmath,amsfonts}
\usepackage{setspace}
\usepackage{textcomp}
\usepackage{stfloats}
\usepackage{url}
\usepackage{verbatim}
\usepackage{booktabs}    % 为了使用\toprule, \midrule, \bottomrule
\usepackage{multirow}    % 为了合并行
\usepackage{xcolor}      % 为了使用颜色
\usepackage{hhline}  
\usepackage{arydshln}
\usepackage{makecell}  
\definecolor{Reviewer1Color}{RGB}{220, 0, 0}    % red
\definecolor{Reviewer2Color}{RGB}{0, 0, 200}    % blue
\definecolor{Reviewer3Color}{RGB}{0, 150, 0}    % green
\definecolor{Reviewer4Color}{RGB}{255, 140, 0}   % orange

\commtrsetup{
title    = {Beyond conventional vision: RGB-event fusion for robust object detection in dynamic traffic scenarios},
author   = {Zhanwen Liu$^{\rm a, {\rm *}}$, Yujing Sun$^{\rm a}$, Yang Wang$^{\rm a}$ , Nan Yang$^{\rm a}$, Shengbo Eben Li$^{\rm b}$, Xiangmo Zhao$^{\rm a}$},
address  = {
{\tdd{a} School of Information Engineering, Chang'an University, Xi'an, 710000, China} \sep

{\tdd{b} School of Vehicle Mobility \& College of AI, Tsinghua University, Beijing, 100084, China} \sep

\tdd{*} Corresponding author.},
email    = {zwliu@chd.edu.cn},
abstract = {
The dynamic range limitation is intrinsic to conventional RGB cameras, which reduces global contrast and causes the loss of high-frequency details such as textures and edges in complex, dynamic traffic environments (e.g., nighttime driving or tunnel scenes). This deficiency hinders the extraction of discriminative features and degrades the performance of frame-based traffic object detection.
 To address this problem, we introduce a bio-inspired event camera integrated with an RGB camera to complement high dynamic range information, and propose a motion cue fusion network (MCFNet), an innovative fusion network that optimally achieves spatiotemporal alignment and develops an adaptive strategy for cross-modal feature fusion, to overcome performance degradation under challenging lighting conditions.
 Specifically, we design an event correction module (ECM) that temporally aligns asynchronous event streams with their corresponding image frames through optical-flow-based warping. The ECM is jointly optimized with the downstream object detection network to learn task-ware event representations.
Subsequently, the event dynamic upsampling module (EDUM) enhances the spatial resolution of event frames to align its distribution with the structures of image pixels, achieving precise spatiotemporal alignment.
Finally, the cross-modal mamba fusion module (CMM) employs adaptive feature fusion through a novel cross-modal interlaced scanning mechanism, effectively integrating complementary information for robust detection performance.
Experiments conducted on the DSEC-Det and PKU-DAVIS-SOD datasets demonstrate that MCFNet significantly outperforms existing methods in various poor lighting and fast moving traffic scenarios. Notably, on the DSEC-Det dataset, MCFNet achieves a remarkable improvement, surpassing the best existing methods by 7.4\% in mAP50 and 1.7\% in mAP metrics, respectively. The code is available at https://github.com/Charm11492/MCFNet.
},
keywords = {
dynamic traffic environments, object detection, multimodal fusion, event camera
},
}

\begin{document}

\maketitle  % 生成标题页

\section{Introduction}
\begin{figure}[!t]
    \centering
    \includegraphics[width=1\columnwidth]{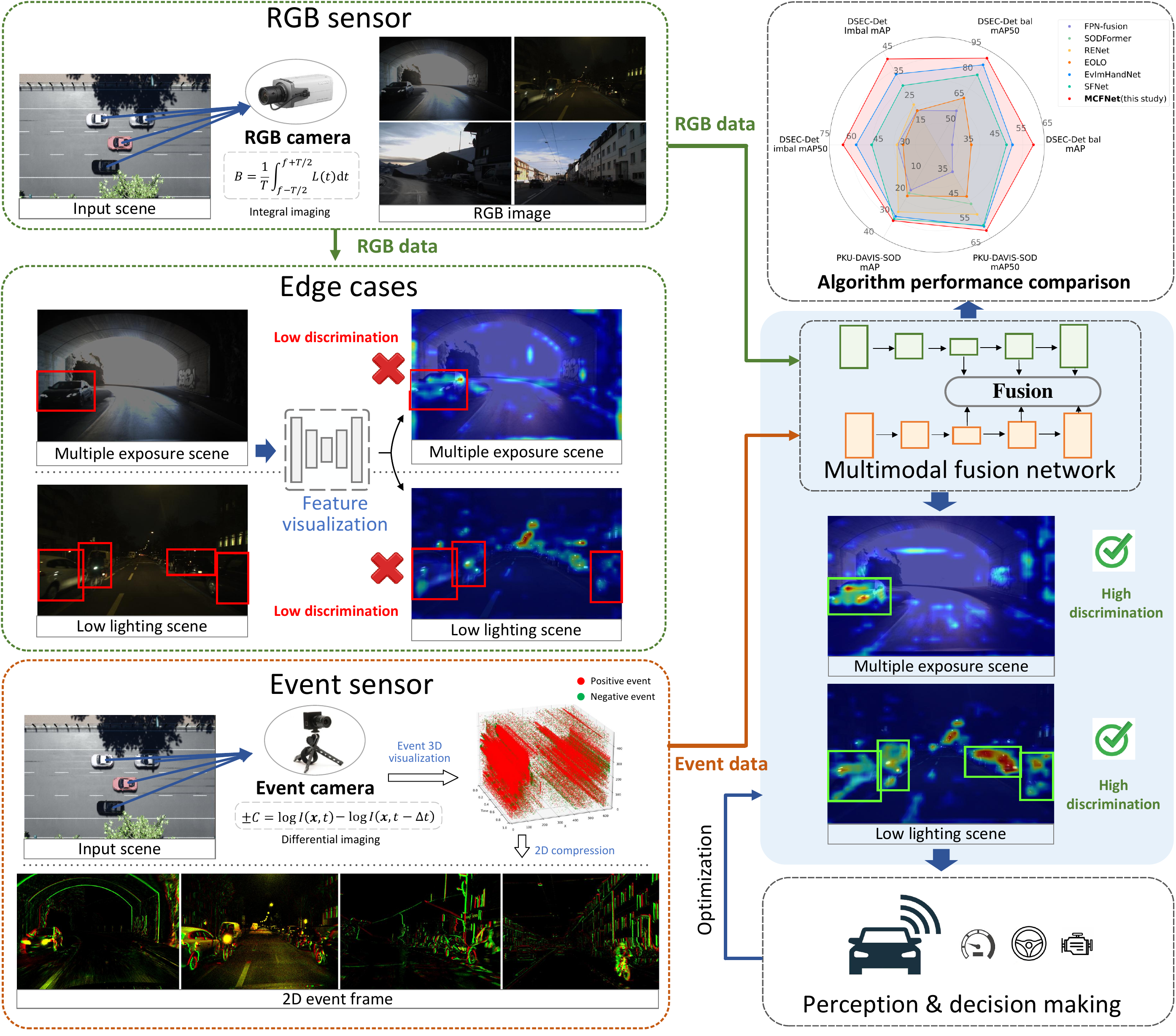}
    \caption{Conventional RGB cameras in boundary perception scenarios are limited by their photosensitive elements and fixed frame rates, making them susceptible to the loss of critical information due to insufficient exposure (upper left), which leads to the failure of downstream algorithms. In contrast, event cameras can capture high-dynamic spatiotemporal features even under extreme conditions (lower left). To this end, we propose a multi-modal fusion algorithm that jointly optimizes the complementarity of RGB and event data to construct a cross-modal feature representation and decision-making mechanism (lower right), thereby significantly enhancing the robustness and accuracy of perception for intelligent vehicles in complex scenarios (upper right).}
    \label{fig_1}
\end{figure}

% \IEEEPARstart{T}{his} file is intended to serve as a ``sample article file''
% for IEEE journal papers produced under \LaTeX\ using
% IEEEtran.cls version 1.8b and later. The most common elements are covered in the simplified and updated instructions in ``New\_IEEEtran\_how-to.pdf''. For less common elements you can refer back to the original ``IEEEtran\_HOWTO.pdf''. It is assumed that the reader has a basic working knowledge of \LaTeX. Those who are new to \LaTeX \ are encouraged to read Tobias Oetiker's ``The Not So Short Introduction to \LaTeX ,'' available at: \url{http://tug.ctan.org/info/lshort/english/lshort.pdf} which provides an overview of working with \LaTeX.
Visual perception systems serve as the foundational component enabling intelligent vehicles to perceive surrounding environments and facilitate decision-making \citep{huang2024human, liao2024gpt, ma2024review}. These systems must maintain robust performance for object detection in complex, dynamic edge cases such as nighttime driving and tunnel navigation. However, conventional RGB cameras, limited by the narrow dynamic range of their intrinsic photosensitive elements, often struggle to capture critical edge details and rapid scene variations under such challenging conditions. As shown in Fig. \ref{fig_1}, this results in degraded image quality, impairing discriminative feature extraction and adversely affecting the accuracy of frame-based traffic object detection methods \citep{berman2011sensors}.

In contrast, bio-inspired event cameras provide a high dynamic range and microsecond-level temporal resolution, allowing for stable imaging under extreme conditions such as low light, and overexposure \citep{chen2024motion,chen2023esvio,rizzo2024exploration}. By harnessing the complementary strengths of RGB and event-based information, multi-modal fusion approaches offer a promising avenue for visual perception in complex traffic environments \citep{bayoudh2022survey}. Recently, numerous object detection algorithms have been developed to exploit RGB and event fusion, significantly enhancing the robustness of visual perception systems in challenging scenarios \citep{cao2024chasing,li2023sodformer,tomy2022fusing}. These methods typically involve designing dedicated multi-modal representation branches to extract features from each modality and implementing advanced fusion modules to deeply integrate cross-modal information, thereby improving detection performance.

However, the heterogeneity between RGB and event cameras causes spatiotemporal inconsistencies in multimodal data, posing challenges for existing methods in multimodal feature extraction and alignment. First, in the temporal dimension, the microsecond-level temporal resolution of event data is significantly higher than the millisecond-level resolution of RGB data, resulting in temporal misalignment and making direct multimodal fusion infeasible \citep{gallego2020event}. To address this issue, existing methods typically perform temporal sampling and compression on event data to generate event frames that align with the temporal resolution of RGB data. These approaches can be broadly categorized into two types: event frame accumulation methods and motion compensation methods. Event frame accumulation methods generate image-like tensors using statistical features of events \citep{kim2021n, maqueda2018event, park2016performance, zhu2018ev, zihao2018unsupervised}, but they are inevitably affected by thermal noise and dark current noise, which degrade the quality of subsequent structural feature extraction. Motion compensation methods exploit the temporal dynamics of event data to align continuous motion events, thereby enhancing motion regions and forming high-intensity edges while suppressing noise with discontinuous motion. However, these methods \citep{Paredes-Vallés_Hagenaars_Croon_2021, paredes2023taming} rely on assumptions of constant illumination and linear motion, which are often violated in highly dynamic traffic scenarios with frequent brightness variations (e.g., vehicle headlights, streetlights) and complex motion patterns, leading to inaccurate motion estimation and degraded event frame quality \citep{basha2013multi, brox2010large}. Second, in the spatial dimension, RGB data typically have much higher resolution than event data, leading to significant significant disparity \citep{gehrig2021dsec}. Existing methods often address this by directly downsampling RGB data to match the resolution of event data, which further exacerbates information loss \citep{tomy2022fusing, zhou2023rgb}.
After extracting representations from both modalities, the spatial inconsistency of illumination distribution results in different discriminability levels of the two modalities across different regions. This necessitates dynamically balancing the contributions of each modality to achieve robust cross-modal feature fusion. Existing methods typically fuse features of the two modalities through simple addition or concatenation \citep{cao2021fusion, tomy2022fusing, yang2025smamba}, which cannot adaptively adjust the fusion ratio based on dynamic scene variations. More recent approaches employ attention mechanisms to capture complementary information \citep{cao2024chasing, li2023sodformer,liu2024enhancing, zhou2023rgb}. However, they either treat one modality as auxiliary to enhance the other or build separate enhancement submodules for each modality, failing to fully exploit inter-modal collaborative information and hinder precise balancing of modality contributions.

To address the aforementioned challenges, we propose a novel motion cue fusion network (MCFNet), which consists of three main components: the event correction module (ECM) and the event dynamic upsampling module (EDUM) for spatiotemporal alignment between RGB and event data to generate high quality pixel to pixel alignment features, and the cross-modal mamba fusion module (CMM) for global contextual modeling across modalities to guide adaptive fusion. Specifically, ECM estimates the motion vectors of the event stream to temporally align the motion events with the RGB frame timestamps. To overcome the limitations of constant illumination and linear motion, an end-to-end joint training strategy that leverages high-level semantic information to guide model optimization is employed. This enables the learning of scene features and the generation of event frames with high-intensity edges and robust noise resilience in highly dynamic traffic scenarios. Building upon this, EDUM is introduced to perform dynamic up-sampling by leveraging the spatial distribution of event features and image local smoothness characteristic to suppress noise, yielding high-resolution and high-quality event features. After extracting high-quality spatiotemporally aligned features from both modalities, CMM employs inter-modal global context interactions to perceive scene content and assess each modality's contribution. This guides the adaptive leveraging of either event dynamics or RGB texture features for cross-modal fusion, fully exploiting their complementary characteristics. To validate the effectiveness of MCFNet, we conducted extensive experiments on the DSEC-Det \citep{gehrig2021dsec} and PKU-DAVIS-SOD \citep{li2023sodformer} datasets. As shown in Fig. \ref{fig_1}, MCFNet significantly outperforms the existing methods.
% MCFNet adopts an end-to-end architecture with joint training, effectively guiding ECM optimization using high-level semantic information. This approach enables the model to learn scene features that mitigate the limitations of conventional assumptions, such as constant illumination and linear motion. ECM computes motion vectors from event data to align timestamps, generating event representations with improved recognition quality. 

In summary, the main contributions of this study are as follows:

1) In temporal alignment, an self-supervised learning based event correction module (ECM) is proposed to estimate full-scene optical flow, providing initial values for joint training. Subsequently, ECM is jointly optimized with the detection network to mitigate the influence of constant lighting and linear motion assumption in optical flow, promising high-quality task-aware event representations.

2) In spatial alignment, a novel event dynamic upsampling module(EDUM) is devised to extract RGB feature smoothness properties to suppress noise amplification during event representation up-sampling, guaranteeing more pure structural features.

3) We propose a cross-modal mamba fusion module (CMM) driven by a cross-modal interlaced scanning mechanism. This mechanism facilitates deep inter-modal feature interaction and global context extraction, significantly enhancing the perception and fusion of complementary cross-modal features.

4) We conduct extensive training and testing using state-of-the-art object detection methods on the DSEC-Det and PKU-DAVIS-SOD datasets, demonstrating that our proposed MCFNet significantly surpasses existing approaches in performance.

\section{Related work}
This section first introduces existing object detection algorithms in dynamic traffic scenarios and then introduces the working principle of event cameras and event representation methods.

\subsection{Object detection in dynamic traffic scenarios}
\textbf{RGB \& RGB-ridar fusion.} \cite{wang2024mamba} proposed the Mamba YOLO detector, which enhances the local modeling capabilities of state-space models (SSMs) in traffic object detection by introducing a residual gating (RG) mechanism. This detector utilizes selective two-dimensional (2D) scanning to process high-resolution image features, achieving competitive performance with linear computational complexity. YOLOv11 further improves traffic object detection by enhancing feature extraction and adopting a lightweight design, delivering both high precision and low latency. These improvements enable rapid responses and support safe decision-making in autonomous driving scenarios \citep{khanam2024yolov11}. Although image-based object detection methods provide excellent real-time performance, RGB cameras are susceptible to performance degradation under adverse weather or varying lighting conditions, which can lead to detection failures \citep{liu2020scale,liu2021cascade,liu2024boosting}. To address this limitation, some researchers have introduced radar sensors, which are immune to lighting variations, and proposed their fusion with RGB images to enhance perception robustness in low-light or dynamically lit environments. For example, \cite{li2024rctrans} developed RCTrans, a radar-camera fusion model based on Transformers. It enhances sparse radar point cloud features using a dense radar encoder and progressively localizes targets via a pruning order decoder. Similarly, \cite{wang2023camo} integrated image and point cloud data to design a module that combines occlusion state estimation with optimal appearance feature selection, effectively mitigating occlusion-related challenges in traffic object tracking for autonomous vehicles. However, radar's relatively low scanning frequency (typically 10--20 Hz) can lead to motion information loss or distortion in high-speed scenarios, potentially compromising detection accuracy and threatening the safe and stable operation of autonomous driving systems \citep{liu2023multi}.

\textbf{RGB-event fusion.} Neuromorphic cameras have recently been introduced to overcome the limitations of RGB cameras and radar in capturing motion information. Due to their asynchronous output, event cameras offer exceptionally high temporal resolution and a wide dynamic range, enabling robust performance in complex and dynamic traffic environments. In the context of object detection using fused RGB and event data, existing approaches can be broadly classified into two categories: late fusion and intermediate fusion.
Late fusion methods operate at the decision level. For example, \cite{Chen_2018} applied non-maximum suppression (NMS) to merge detection outputs from both modalities; \cite{li2019event} employ dempster-shafer theory to integrate events and frames for vehicle and pedestrian detection; and the related study \citep{Jiang_Xia_Huang_Stechele_Chen_Bing_Knoll_2019} proposed combining confidence maps derived from each modality. However, these approaches typically suffer from insufficient feature interaction and fail to fully leverage the complementary properties of RGB and event data.
In contrast, intermediate fusion strategies perform integration at the feature level. \cite{tomy2022fusing} adopt a straightforward concatenation of event and frame-based features to enhance performance; SFNet introduces two dedicated sub-modules to refine features from each modality for effective cross-modal fusion \citep{liu2024enhancing}; RENet designs a bidirectional fusion module to capture multi-modal features across spatial and channel dimensions \citep{zhou2023rgb}; SODformer proposes an asynchronous attention-based fusion mechanism \citep{li2023sodformer}; and EOLO incorporates a symmetric fusion module with attention mechanisms to align RGB and event features \citep{cao2024chasing}. EvImHandNet \citep{jiang2024complementing} leverages the complementary advantages of event and RGB cameras for 3 dimensional (3D) hand mesh reconstruction through a fusion module that employs spatial attention mechanisms and fully connected layers for weighted fusion. However, it still suffers from limited cross-modal interaction. To address this issue and enhance cross-modal feature integration, CAFR \citep{cao2024embracing} introduces a self-attention-based fusion module that separately computes {rgb\_cross\_attention} and {event\_cross\_attention}. This dual-path design introduces parameter redundancy, resulting in increased computational overhead. Despite their advantages, these methods predominantly emphasize inter-modal feature interactions while neglecting the joint modeling of global contextual information from both modalities. This oversight limits the network's ability to adaptively exploit dynamic cues from event data and rich textures from RGB images in accordance with varying scene characteristics.

Furthermore, most fusion methods assume that both modalities share the same spatial resolution. However, widely used sensors such as asynchronous time-based image sensors (ATISs) and dynamic and active pixel vision sensors (DAVISs), which output both event streams and intensity images at shared pixel locations, offer relatively low resolution and cannot meet the demands of traffic perception tasks. More commonly, modern imaging systems spatially separate RGB and event cameras, enabling the acquisition of high-resolution RGB images alongside event data. When conducting multimodal fusion on such data, resolution inconsistency must be addressed. Existing solutions often downsample the RGB images to match the resolution of event data, which results in a loss of high-frequency visual information and degraded perception performance. To mitigate this, we propose performing dynamic upsampling of event features prior to fusion, thereby maximizing the use of rich visual content and enhancing fusion effectiveness. 

% \subsection{Event Camera}
% Event camera is a neuromorphic sensor whose photosensitive chips consist of multiple independently operating comparators that generate events by detecting changes in light intensity\cite{gallego2020event}. Specifically, when a comparator at a pixel location perceives a brightness change and the voltage variation exceeds a preset threshold $C$, it generates an event $e_k$, which includes the position $u=(x, y)$, timestamp $t_k$, and polarity $p_k$.
% \begin{equation}
%     \Delta L(u,t_k)=L(u,t_k)-L(u,t_k-\Delta t_k)=P_kC,
% \end{equation}
% where $P_k\in\{-1,1\}$ denotes the polarity of the event, and the values 1 and -1 indicate an increase or decrease in intensity at the pixel, respectively. Where $L(u,t_k)$ and $L(u,t_k-\Delta t_k)$ denote the illumination intensity at time $t$ and $t-\Delta t$, respectively.

% \subsection{RGB and Event fusion based object detection}

\subsection{Event representation}
The event camera is a neuromorphic sensor whose photosensitive chips consist of multiple independently operating comparators that generate events by detecting changes in light intensity \citep{gallego2020event}. Specifically, when a comparator at a pixel location perceives a brightness change and the voltage variation exceeds a preset threshold $C$, it generates an event $e_k$, which includes the position $u=(x, y)$, timestamp $t_k$, and polarity $p_k$. The process is as
\begin{equation}
    \Delta L(u,t_k)=L(u,t_k)-L(u,t_k-\Delta t_k)=P_kC
\end{equation}
where $P_k\in\{-1,1\}$ denotes the polarity of the event, and the values 1 and --1 indicate an increase or decrease in intensity at the pixel, respectively; $L(u,t_k)$ and $L(u,t_k-\Delta t_k)$ denote the illumination intensity at time $t$ and $t-\Delta t$, respectively.

Effectively representing event data is crucial for accurate modeling. Existing fusion frameworks often convert asynchronous events into dense, image-like representations to enable the use of mature machine learning algorithms and neural network architectures. These dense representations can be broadly classified into four categories: image-based methods \citep{maqueda2018event, zhu2018ev}, timestamp-based methods \citep{kim2021n, park2016performance}, voxel-based methods \citep{zihao2018unsupervised}, and motion compensation methods \citep{gallego2018unifying, gu2021spatio, Paredes-Vallés_Hagenaars_Croon_2021, paredes2023taming, shiba2022secrets}.
Image-based methods typically rely on event statistics or polarity counting, which often leads to a significant loss of temporal information. Timestamp-based methods, such as Time Surface \citep{lagorce2016hots}, HATS \citep{sironi2018hats}, and DiST \citep{kim2021n}, apply temporal decay within event windows, assigning higher weights to more recent events. Voxel-based methods, like Voxel Grid \citep{zihao2018unsupervised}, divide raw event streams into temporal bins and apply interpolation techniques to build voxelized representations. However, these methods are highly susceptible to thermal noise and dark current noise, which can degrade the quality of subsequent structural information extraction. 

Recent motion compensation approaches \citep{gallego2018unifying, gu2021spatio, shiba2022secrets, Paredes-Vallés_Hagenaars_Croon_2021, paredes2023taming,10612786} have shown promising results and can be categorized into model-based, self-supervised, and supervised learning methods. Model-based techniques \citep{gallego2018unifying, gu2021spatio, shiba2022secrets} simulate motion and optimize models based on contrast maximization. Self-supervised methods typically estimate optical flow by assuming brightness constancy and linear motion, which are often violated in real-world conditions involving complex motion or varying lighting (e.g., vehicle headlights and streetlights), leading to artifacts and noise in reconstructed frames. To overcome these limitations, some studies \citep{Paredes-Vallés_Hagenaars_Croon_2021, paredes2023taming} incorporate recurrent structures to enhance temporal modeling and mitigate the shortcomings of overly simplistic assumptions. However, these methods often increase the training complexity, and their task-decoupled design limits the optical flow network's ability to model complex motion in semantically rich scenes. Supervised methods, on the other hand, require ground truth optical flow, which is often lacking in most object detection datasets and impractical for real-world applications, thus not discussed here.

\section{Proposed method}
This section first introduces the principle of Mamba in Section \ref{tab:Mamba}, then describes the overall architecture of the proposed method in Section \ref{tab:MCFNet}. Section \ref{tab:ECM} introduces the detailed information of the ECM, revealing its internal working principles. Section \ref{tab:EDUM} presents the design of the EDUM. Finally, Section \ref{tab:CMM} introduces the CMM.
% \subsection{Event Camera}
% \label{tab:Event}
% Event camera is a neuromorphic sensor whose photosensitive chips consist of multiple independently operating comparators that generate events by detecting changes in light intensity\cite{gallego2020event}. Specifically, when a comparator at a pixel location perceives a brightness change and the voltage variation exceeds a preset threshold $C$, it generates an event $e_k$, which includes the position $u=(x, y)$, timestamp $t_k$, and polarity $p_k$.
% \begin{equation}
%     \Delta L(u,t_k)=L(u,t_k)-L(u,t_k-\Delta t_k)=P_kC,
% \end{equation}
% where $P_k\in\{-1,1\}$ denotes the polarity of the event, and the values 1 and -1 indicate an increase or decrease in intensity at the pixel, respectively. Where $L(u,t_k)$ and $L(u,t_k-\Delta t_k)$ denote the illumination intensity at time $t$ and $t-\Delta t$, respectively.

\subsection{Introduction of SSMs and mamba}
\label{tab:Mamba}
Mamba represents a novel sequence modeling architecture, fundamentally based on selective SSMs \citep{gu2023mamba}. SSM is a linear model designed to characterize the dynamic behavior of systems over time, mapping 1 dimensional (1D) input signals $x(t)\in R$ to $N$- dimensional latent states $h(t)\in R^N$, before projecting them back to 1D output signals $y(t)$. The mathematical formulation is expressed as
\begin{equation}
    \left.\left\{
\begin{aligned}
h^{\prime}(t) & =Ah(t)+Bx(t) \\
y(t) & =Ch(t)+Dx(t)
\end{aligned}\right.\right.
\label{eq:2}
\end{equation}
where $A\in R^{N\times N}$ is the state transition matrix, $B\in R^{N\times 1}$ is the mapping matrix from input to state, $C\in R^{N\times 1}$ is the state-to-output matrix, and $D\in R$ is the input-to-output parameter.

Mamba introduces the selective state space scanning (S6) mechanism, which incorporates a learnable time-scale parameter $\Delta$ that dynamically adapts to the input \citep{gu2023mamba}. This design overcomes the linear time-invariant (LTI) limitation of traditional SSMs and enhances global modeling capability. Specifically, $\Delta$ controls the discrete-time dynamics by scaling the state transition at each input position, enabling the model to adjust how quickly or slowly it updates its internal state. As a result, the model can respond rapidly to abrupt changes with smaller $\Delta$, or integrate information over longer contexts with larger $\Delta$, achieving fine-grained control over temporal resolution. The model is discretized using a zero-order hold (ZOH), as detailed below:
\begin{equation}
\bar{A}=\exp(\Delta A), \quad \bar{B}=(\Delta A)^{-1}(\exp(\Delta A)-I)\Delta B
\label{eq:3}
\end{equation}
\begin{equation}
h_{t}=\overline{A} h_{t-1}+\overline{B} x_{t}, \quad y_{t}=C h_{t}+ D x_{t}
\label{eq2}
\end{equation}
where $\Delta A$ and $\Delta B$ represent the product of matrices $A$ and $B$ respectively with the scalar $\Delta$, which plays the role of scaling continuous-time state changes and inputs; $\bar{A}$ is the discrete state transfer matrix, $\bar{B}$ is the discretized input matrix, and $I$ is the identity matrix. In Eq.~(\ref{eq:2}), $h_t$ denotes the hidden state at discrete time step $t$, and $x_t$, $y_t$ are the input and output, respectively. Our proposed CMM module is built upon this mamba architecture.

\subsection{Overview of MCFNet}
\label{tab:MCFNet}

\begin{figure*}[h]
\makebox[\textwidth][c]{
    \includegraphics[width=6.5in]{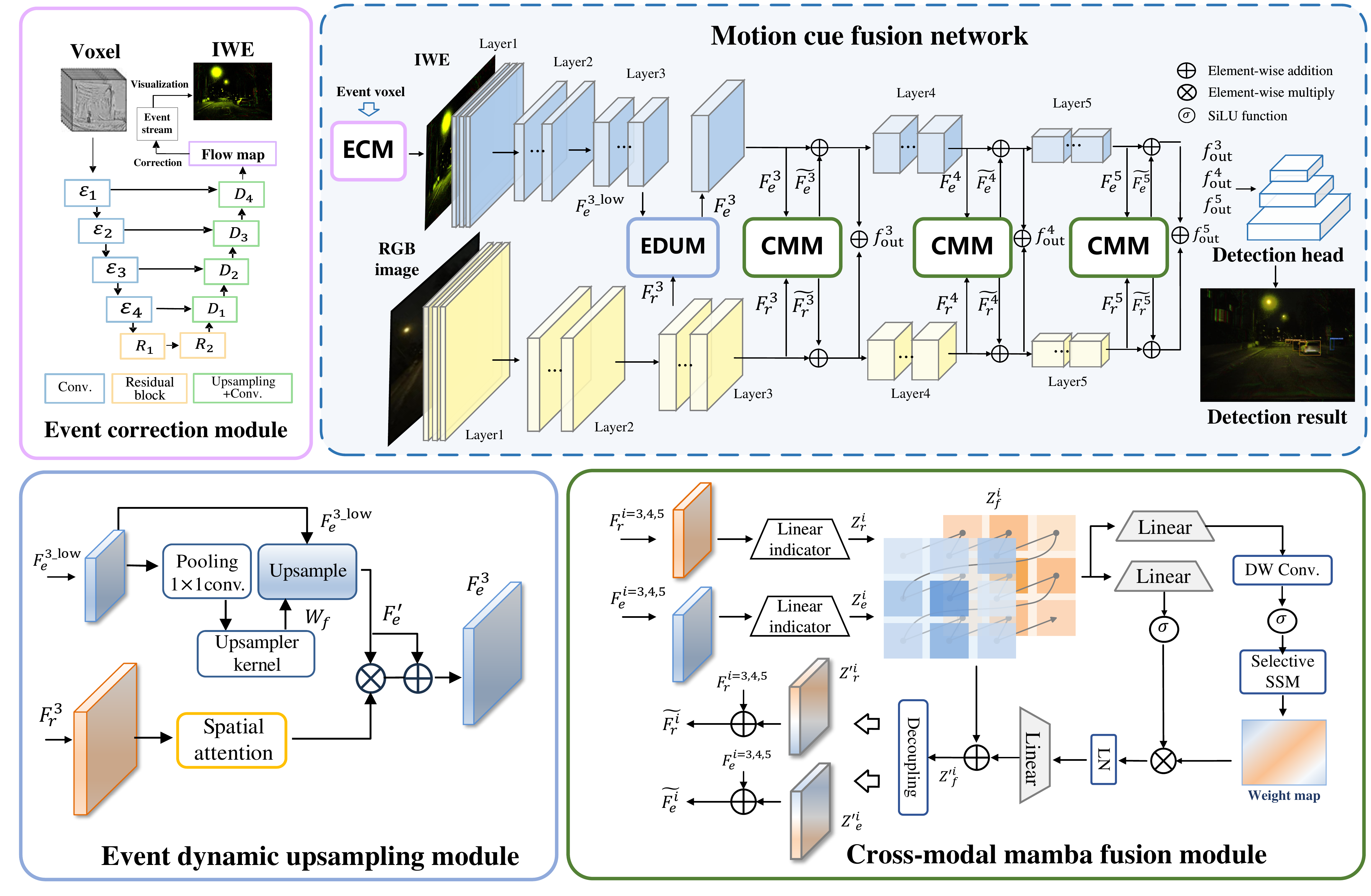}
}

\caption{Pipeline of the proposed motion cue fusion network (MCFNet) is as follows. Voxels generated from raw event streams are first processed by the event correction module (ECM; see Section~\ref{tab:ECM}) to produce high-quality event frames. These frames, along with RGB images, are then fed into two separate CSPDarkNets for modality-specific feature extraction. The event dynamic upsampling module (EDUM; see Section~\ref{tab:EDUM}) takes the features from the stage-3 layer to align the spatial resolutions of both modalities. This is followed by the cross-modal mamba fusion module (CMM; see Section~\ref{tab:CMM}) to perform cross-modal fusion. FPN combined with PANet further integrates multi-scale features, and finally, the decoder outputs category and bounding box predictions for each detected object.}
\label{fig_2}
\end{figure*}

In this section, we present the motion cue fusion network (MCFNet), as illustrated in Fig.~\ref{fig_2}, which takes RGB images and corresponding event streams as inputs, and achieve precise pixel-level spatiotemporal alignment and adaptive feature fusion across modalities. As shown in Fig.~\ref{fig_2}, the ECM estimates optical flow fields to temporally align event streams through warping operations, generating high-quality event representations tailored for detection. Subsequently, two parallel branches are used for modality-specific feature extraction, independently processing RGB images and event frames. Each branch adopts the CSPDarkNet backbone from the YOLOX framework \citep{ge2021yolox}, producing multi-scale feature maps. To address the resolution mismatch between RGB and event sensors in modern imaging systems, we introduce the EDUM, which adaptively refines upsampling kernels based on the spatial distribution of event features. This design facilitates the generation of high-resolution, high-fidelity event representations. Next, the CMM fusion module fuses features from both modalities at three hierarchical levels, adaptively integrating complementary information while preserving global contextual cues. Finally, the fused multi-scale features are processed by the FPN+PANet framework \citep{Lin_Dollar_Girshick_He_Hariharan_Belongie_2017, liu2018path}, followed by a detection head based on YOLOX for object classification and bounding box regression.

\subsection{Event correction module}
\label{tab:ECM}
Due to the difference in temporal resolution between RGB and event cameras, there exists a temporal misalignment between the two modalities, which adversely affects precise cross-modal fusion and the extraction of discriminative features. To address this issue, motion compensation methods estimate the optical flow of the event stream to map events to the timestamp of the corresponding RGB frame, thereby achieving temporal alignment across modalities. However, these methods rely on assumptions of illumination consistency and linear motion, which are often violated in highly dynamic traffic scenarios, leading to inaccurate optical flow estimation and temporal alignment. To overcome the limitations, we propose an end-to-end architecture that integrates an optical flow-based ECM with an object detection network. This architecture facilitates the joint optimization of optical flow estimation and object detection tasks. By feeding the output of the ECM into the detection network and utilizing a unified backpropagation mechanism, the object detection supervisory signals can directly guide the learning of optical flow, thereby enabling the model to capture scene features that go beyond conventional assumptions of constant illumination and linear motion. As a result, the generated event representations are better tailored to the requirements of object detection.

Specifically, we extract the event stream $e = {(x_k, y_k, t_k, p_k)}{k=1}^N$ between the exposures of two consecutive image frames, where $t_0 < t_k < t_0 + \Delta t$, and to be consistent with the RGB sampling interval, $\Delta t$ is set to 50 ms to define the temporal window. To convert the unstructured event stream into a tensor form acceptable to the network, we adopt a voxel-based event representation method \citep{zihao2018unsupervised}. Specifically, we map the event stream $e$ into a three-dimensional voxel grid $\mathrm{E} \in \mathbb{R}^{B \times H \times W}$, where $B$ is the number of temporal bins. Following the setting in \cite{zihao2018unsupervised}, we set $B = 5$. The events are assigned to bins based on their timestamps, and we first normalize the event timestamps accordingly:
\begin{equation}
    t_{k}^{*}=\frac{t_{k}-t_{0}}{t_{-1}-t_{0}},   t_{k}^{*} \in[0,1]
\end{equation}
where $t_0$ and $t_{-1}$ represent the timestamps of the first and last events in the current event sequence, respectively. For each partition, each event (indexed as $k$) is assigned its polarity $p_k$ to the two nearest time bins according to the following formula, with the weights determined by the triangular kernel function:
\begin{equation}
    E(x_k,y_k, t) = \sum_k p_i \times \kappa\left(t - t_k^*(B - 1)\right),  \kappa(a) = \max(0, 1 - |a|)
\end{equation}

ECM predicts horizontal and vertical optical flow for each pixel within the interval and warps the events to align with a reference timestamp $t_{\textrm{ref}}$, achieving temporal alignment across modalities:
\begin{equation}
\begin{aligned}
x_k^\prime &= x_k + (t_{\text{ref}} - t_k) \times u(x_k, y_k) \\
y_k^\prime &= y_k + (t_{\text{ref}} - t_k) \times v(x_k, y_k)
\end{aligned}
\end{equation}

We then apply bilinear interpolation to the compensated event stream $e' = {(x_k^\prime, y_k^\prime, t_k, p_k)}_{k=1}^N$ to map events to their nearest pixel locations, using the surrounding four neighbors and performing a weighted sum based on polarity and visualize it, finally obtaining the $\text{Image of Warping Event (IWE)} \in \mathbb{R}^{ H \times W \times 3}$:
\begin{equation} \text{IWE} = \sum_k p_k \cdot \max\left(0, 1 - \left|e'_{k}{(x,y)} - e_{k}{(x,y)}\right|\right) \end{equation}

\textbf{Loss function.} In ECM, we choose to train in a self-supervised manner and use a contrast maximization framework loss for motion compensation. Following the approach in \cite{9341224}, we use bilinear interpolation to generate an average timestamp image for each pixel under each polarity $p^{\prime}$. And we optimize the loss function by minimizing the sum of squared differences between the forward- and backward-warped event images:
\begin{equation}
\begin{aligned}
T_p^*(x;(u,v)|t_{\text{ref}}) = &\frac{\sum_j \kappa(x - x_j')\kappa(y - y_j')t_j^*}
{\sum_j \kappa(x - x_j')\kappa(y - y_j') + \epsilon}\\
j = \{ i \mid p_i = &p' \}, \quad p' \in \{+,-\}, \quad \epsilon \to 0
\end{aligned}
\end{equation}
where $(u, v)$ is the estimated optical flow, $(x_j', y_j')$ are the warped event coordinates, $t_j^*$ is the normalized timestamp, and $\kappa(\cdot)$ denotes the bilinear interpolation kernel. The contrast loss is then defined as
\begin{equation}
\begin{aligned}
    \mathcal{L}_{\text{contrast}}(t_{\text{ref}}^*) &= \sum_x \left[ T_+(x;(u,v)|t_{\text{ref}}^*)^2 + T_-(x;(u,v)|t_{\text{ref}}^*)^2 \right] \\
    \mathcal{L}_{\text{contrast}} &= \mathcal{L}_{\text{contrast}}(1) + \mathcal{L}_{\text{contrast}}(0)
\end{aligned}
\end{equation}

% \textcolor{red}{In addition, a smoothness loss is introduced to regularize the estimated optical flow by minimizing the flow differences between neighboring pixels \cite{zhu2018ev}. Here, $N$ denotes the set of neighboring pixels around a given pixel $(x,y)$. }
To regularize the optical flow estimation, we introduce a smoothness loss that penalizes discontinuities in the flow field \citep{zhu2018ev}. For each pixel $(x,y)$, its neighbors are denoted by $\mathcal{N}(x,y)$, and the robust Charbonnier penalty function $\rho(\cdot)$ is applied to flow gradients:
\begin{equation}
\begin{aligned}
    l_{\text{smoothness}} = \sum_{x,y} \sum_{i,j \in \mathcal{N}(x,y)} \rho(u(x,y) & - u(i,j)) + \rho(v(x,y) - v(i,j))\\
    \rho(x) = & \sqrt{x^2 + \varepsilon^2}
\end{aligned}
\end{equation}
The total loss is a weighted sum of contrast loss and smoothness loss:
\begin{equation}
    \mathcal{L}_{\text{ECM}} = \mathcal{L}_{\text{contrast}} + \lambda_1 \mathcal{L}_{\text{smooth}}
\end{equation}

\textbf{Training strategy.} Benefiting from the fast convergence and strong generalization ability of ECM, we first jointly train the network on a representative subset of the DSEC-Det dataset. The trained model is then used to generate deterministic IWEs for the entire dataset, which serve as input to train the object detection component of MCFNet.  This strategy ensures high detection performance while significantly reducing training costs. To initialize the network, we use the official pre-trained weights from \cite{zhu2018ev} and \cite{ge2021yolox}.

\subsection{Event dynamic upsampling module}
\label{tab:EDUM}
Due to differences in resolution and field of view between RGB and event cameras, their outputs exhibit spatial misalignment. To compensate for this spatial discrepancy, existing fusion methods often downsample RGB frames to match the resolution of event data at the input stage \citep{liu2024enhancing, tomy2022fusing, zhou2023rgb}, which inevitably leads to the loss of high-frequency details and compression of informative content in RGB images, ultimately degrading perception performance. To address this issue, we propose the EDUM, which upsamples event features prior to modal fusion to fully exploit high-quality RGB information. However, conventional upsampling approaches typically rely on fixed-parameter transposed convolutions to enhance spatial resolution, ignoring variations in pixel distributions across different scenes. This often results in blurred details or visual artifacts. In contrast, our EDUM adaptively modulates upsampling weights based on the input features, allowing for more fine-grained and context-aware resolution enhancement.

Specifically, EDUM first applies global average pooling to the input event features $F_{e}^{3 \_\textrm{low}}\in R^{B\times C\times H\times W}$ and implements sequential interaction through $1\times1$ convolution. The learned weights are then assigned to the transposed convolution kernel $W\in R^{B\times C_{\textrm{in}}\times C_{\textrm{out}}\times3\times3}$ to perform transposed convolution upsampling on low-resolution Event features, yielding high-resolution Event features $F_{e}^{\prime}\in R^{B\times C\times2H\times2W}$. The formulation is as
\begin{equation}
    D=GAP(F_{e}^{3 \_\textrm{low}})
\end{equation}
\begin{equation}
    W_f=W\otimes\mathrm{Con}v_{1\times1}(D)
\end{equation}
\begin{equation}
    F_e^{\prime}=\text{DeConv}(W_f,F_{e}^{3 \_\textrm{low}})
\end{equation}
where $GAP$ represents the global average pooling and $\text{DeConv}$ represents the de-convolution operation.

Simultaneously, considering that event cameras inevitably generate noise due to their sensitivity to dark current and photocurrent, we leverage spatial attention maps derived from high-resolution RGB features of the third layer of the backbone network $F_{r}^{3}\in R^{B\times C\times2H\times2W}$ to suppress noise amplification during upsampling.  We exploit the smoothness of RGB features to enhance the up-sampled event features, ultimately obtaining enhanced high-resolution event features $F_{e}^{3}\in R^{B\times C\times2H\times2W}$. The formulation is as
\begin{equation}
F_e^{3}=F_e^{\prime}+F_e^{\prime}\otimes\sigma(\text{Concat}(\text{AveP}(F_{r}^{3}),\textrm{MaxP}(F_{r}^{3}))
\end{equation}
where $\text{AveP}$ is an average pooling layer, $\textrm{MaxP}$ is a max pooling layer, and $\sigma$ is the sigmoid function.

\subsection{Cross-modal mamba fusion module}
\label{tab:CMM}

To simultaneously perceive the most valuable complementary information during cross-modal interaction, and adaptively leverage Event dynamics or RGB texture features for complementary fusion based on scene characteristics, we design CMM with a cross-modal interlaced scanning mechanism. This mechanism simultaneously performs inter-modal feature interaction and captures global information, selectively preserving salient features while filtering redundant information.

Specifically, we choose to fuse the two modalities in the third, fourth, and fifth layers of the backbone network and input the fused features into FPN+PANet. Therefore, we first project the two input modality features of CMM, $F_{e} ^{i=3,4,5}\in R^{C\times H\times W}$ and $F_{r}^{i=3,4,5}\in R^{C\times H\times W}$, to the latent space through a linear transformation, where \textit{i} represents the feature from the \textit{i}-th layer of the backbone network. The magnitude and offset of modal features are then adjusted through scaling factors to accommodate different modal feature distributions, enabling feature domain alignment and distribution normalization across different modalities.
\begin{equation}
\left\{
\begin{aligned}
    Z_e^{i} &= F_e^{i} \odot r_e^{i} + \beta_e^{i} \\
    Z_r^{i} &= F_r^{i} \odot r_r^{i} + \beta_r^{i}
\end{aligned}
\right.
\end{equation}
where $r_e^{i}/r_r^{i}$ and $\beta_e^{i}/\beta_r^{i}$ represent the scaling and offset factors respectively, and $\odot$ denotes the element-wise multiplication operation. 

Subsequently, $Z_e^{i}$ and $Z_r^{i}$ are concatenated along the $W$ dimension to obtain the feature-level fine-grained fusion features $Z_f^{i} \in \mathbb{R}^{C\times H\times2W}$. This concatenation operation provides the spatial and structural foundation for fusion, enabling the interaction of information from both modalities within a unified representation space.
\begin{equation}
Z_f^{i}=\textrm{Cross\_concat}(Z_e^{i},Z_r^{i},\mathrm{dim}=W)
\end{equation}

The mixed features $Z_f^{i}$ are flattened along the $H$ and $W$ dimensions and then fed into the $SSMs$, allowing each modality to selectively extract relevant features based on information from the other. Through global spatial interactions between the two modalities, a WeightMap is generated to reflect their complementary relationship, guiding the adaptive utilization of either the dynamic properties of events or the rich semantic features of RGB. Additionally, the integration of depth-wise separable convolution (DWConv) and Layer Normalization further improves computational efficiency while enhancing the capability to model cross-modal correlations. As a result, enhanced mixed features ${Z^{\prime}}_f^{i}$ are generated. The formulation is as
\begin{equation}
\textrm{WeightMap} = \mathcal{\mathrm{SSM}}(\overline{\mathbf{\textit{A}}}, \overline{\mathbf{\textit{B}}}, \textit{C})(\mathbf{Z^{\prime}}^{i}_{f} )
\end{equation}
\begin{equation}
    {Z^{\prime}}_f^{i} =\text{LN}(\text{Liner}(\textrm{WeightMap}\odot Z_f ))
\end{equation}
where $\overline{\textit{A}}$, $\overline{\textit{B}}$, and $\textit{C}$ are defined in Eqs. (\ref{eq:2}) and (\ref{eq:3}), and represent the discretized state transition matrix, the input-to-state mapping matrix, and the state-to-output matrix, respectively.
Finally, the feature ${Z^{\prime}}_f^{i}$ is decoupled back to the two modalities, generating enhanced modal features ${Z^{\prime}}_r^{i}$ and ${Z^{\prime}}_e^{i}$, which are then fused with the original features $Z_e^{i}$ and $Z_r^{i}$ through residual connections to obtain $\widetilde{F^{i}}_e$ and $\widetilde{F^{i}}_r$, respectively. This approach effectively enhances modality complementarity while preserving the original feature information. The formulation is as
\begin{equation}
    {Z^{\prime}}_e^{i},{Z^{\prime}}_r^{i}=\text{decouple}({Z^{\prime}}_f^{i})
\end{equation}
\begin{equation}
    \widetilde{F^{i}}_e=F_e^{i}+{Z^{\prime}}_e^{i}
\end{equation}
\begin{equation}
    \widetilde{F^{i}}_r=F_r^{i}+{Z^{\prime}}_r^{i}
\end{equation}

We then add and fuse the adaptively enhanced features from the two modalities to obtain $f^{i}_{\text{out}}$, which is subsequently fed into the FPN+PANet for multi-scale semantic enhancement and feature integration.

\section{Experiments}
This section first describes the used dataset as well as the experimental settings. Subsequently, quantitative and qualitative results are presented to demonstrate the effectiveness of our method. Finally, we perform an ablation study for each module in our network.

\subsection{Datasets and experimental settings}
In the experiments, we verify the effectiveness of our model on two real scene datasets, namely DSEC-Det \citep{gehrig2021dsec} and PKU-DAVIS-SOD \citep{li2023sodformer}. 

\textbf{DSEC-Det.} The dataset contains wide-baseline stereo data captured using an RGB camera (1440×1080) and a high-resolution monochrome event camera (640×480). It comprises 53 sequences, with 39 used for training and 14 for testing, totaling 6.39×$10^4$ frames \citep{gehrig2021dsec}.

The DSEC-Det dataset has multiple versions of annotations. Among them, the automatic annotation methods used by DAGr \citep{gehrig2024low} and FPN-fusion \citep{tomy2022fusing} lead to degraded labeling quality, while SFNet \citep{liu2024enhancing} provides manually annotated labels that are more comprehensive and accurate. For this reason, we adopt the annotation information from SFNet \citep{liu2024enhancing}, and follow their data split strategy to maintain fair comparisons.

\textbf{PKU-DAVIS-SOD.} The PKU-DAVIS-SOD dataset collected by the DAVIS346 event camera provides spatially aligned RGB frame and event stream (346×260) data, with 6.713×$10^5$ labels for training and 2.141×$10^5$ for testing. Each subset is further divided into three typical scenes (normal, motion blur, and low-light) \citep{li2023sodformer}.

\textbf{Implementation details.} We train the MCFNet using the Adam optimizer with a learning rate of 5×$\textrm{e}^{-4}$ and a batch size of 2. The network architecture incorporated CSPDarkNet from YOLOX \citep{ge2021yolox} as both the RGB backbone and Event backbone. The training pipeline implemented standard YOLO data augmentation methods for RGB input, including mosaic and mixup techniques.

Due to the inconsistent resolution of the two modalities in the two datasets, different configurations are adopted. For experiments on the PKU-DAVIS-SOD dataset, where RGB and Event data share the same resolution, our method is implemented without the EDUM module. In comparative experiments, since EvImHandNet \citep{jiang2024complementing} is originally designed for hand mesh reconstruction, we only replace our CMM with its fusion module CFM for fair comparison while retaining all other architectural components. For other methods, we follow their original training strategies to ensure a fair comparison.

\textbf{Evaluation metrics.} We adopt COCO metrics \citep{Lin_Maire_Belongie_Hays_Perona_Ramanan_Dollár_Zitnick_2014} to evaluate the accuracy of object detection, including mAP50 with an IOU threshold of 50\% and mAP, which averages over IOUs between 50\% and 95\%. To assess model efficiency, we also report parameters, FLOPs, and runtime.

\subsection{Quantitative results}
To evaluate the superiority of our approach, we compare MCFNet with state-of-the-art (SOTA) object detection methods. Furthermore, to validate the effectiveness of our event representation strategy, we conduct comparative experiments against leading event representation and motion compensation techniques.

\textbf{Comparison with SOTA object detection methods. } We compare our method with recent SOTA object detection methods, including two event-based methods: RVT \citep{gehrig2023recurrent} and SAST \citep{peng2024scene}; three RGB-based methods: YOLOv11 \citep{khanam2024yolov11}, YOLOv12 \citep{tian2025yolov12} and Mamba-YOLO \citep{wang2024mamba}; and four fusion-based methods: FPN-fusion \citep{tomy2022fusing}, SODFormer \citep{li2023sodformer}, EOLO \citep{cao2024chasing}, and SFNet \citep{liu2024enhancing}. In addition, to compare with other fusion methods, we replace our CMM with the CNN-based fusion method CFM in EvImHandNet \citep{jiang2024complementing} and the transformer-based fusion module in CAFR \citep{cao2024embracing}.

% RENet \cite{zhou2023rgb}
As shown in Table \ref{tab:SOTA}, our MCFNet outperforms existing object detection methods on both datasets. Specifically, on the class-imbalanced DSEC-Det dataset, our method outperforms the second-best method by 1.7\% and 7.4\% in mAP and mAP50, respectively. On the class-balanced DSEC-Det dataset, our method shows consistent superiority, with mAP and mAP50 improved by 1.7\% and 3.3\%, respectively. Furthermore, when comparing different methods, although our MCFNet achieves leading performance in detection accuracy, it incurs a higher computational cost, with 227.8 G FLOPs and an inference time of 47.3 ms, compared to lightweight unimodal models such as Mamba-YOLO (49.6G) \citep{wang2024mamba} and YOLOv11 (35.7 ms) \citep{khanam2024yolov11}. Nevertheless, MCFNet maintains high accuracy while achieving an inference speed of approximately 21 FPS, demonstrating preliminary real-time processing capability. Furthermore, compared to other multimodal methods (e.g., SFNet, which has a similar inference time of 44.8 ms but significantly lower accuracy), our approach presents a favorable balance between performance and efficiency.

\begin{table*}[h]
\belowrulesep=0pt
\aboverulesep=0pt
\centering
\caption{Performance comparison of SOTA object detection methods under different datasets (mAP50/ mAP). The best and the second-best performances are marked in red bold and blue bold, respectively. Note: M denotes million parameters.}
\label{tab:SOTA}
\renewcommand\arraystretch{1.5}
\label{tab:comparison}
\scalebox{0.8}{
% \begin{tabular}{c|cc|cccc|ccc|c}
\begin{tabular}{ccccccccccc}
\Xhline{2pt}
& \multirow{2}{*}{Method} & \multirow{2}{*}{Pub. \& Year
} & \multicolumn{4}{c}{DSEC-Det} & \multicolumn{3}{c}{PKU-DAVIS-SOD} & \multirow{2}{*}{Parameter} \\
\cmidrule(lr){4-6} \cmidrule(lr){7-10}
% \cline{4-10}
& & & \makecell{Class-balanced\\mAP50/mAP} & \makecell{Class-imbalanced\\mAP50/mAP} & FLOPs & Runtime &  \makecell{mAP50/\\mAP} &  {FLOPs} & {Runtime} \\
\hline
\multirow{3}{*}{\rotatebox{90}{RGB}} & YOLOv11 & arXiv'24 & 84.5/\textcolor{blue}{59.2} & 59.1/\textcolor{blue}{40.1} & 102.3 G & 35.7 ms & 58.0/30.9 & 44.2 G & 15.6 ms & 20.1 M \\
& Mamba-YOLO & AAAI'25 & 83.4/55.6 & 53.9/34.7 & 49.6 G & 58.7 ms & 57.1/29.8 & 21.6 G & 21.0 ms & 21.8 M \\
& YOLOv12 & arXiv'25 & 86.8/58.8 & 55.0/34.6 & 90.1 G & 46.5 ms & 60.2/32.4 & 12.22 G & 13.4 ms & 19.6 M \\
\hline
\multirow{2}{*}{\rotatebox{90}{Event}} & RVT &CVPR'23 & 51.1/26.6 & 25.1/12.9 & 19.6 G & 11.9 ms & 50.3/25.6 &6.5 G &7.1 ms &18.5 M \\
& SAST &CVPR'24 & 53.8/39.3 & 24.3/12.1 & 18.5 G & 18.8 ms & 48.7/24.5 &6.2 G &16.7 ms &18.5 M \\
\hline
\multirow{8}{*}{\rotatebox{90}{RGB-Event}}
& FPN-fusion &ICRA'22 & 56.8/30.7 & 36.9/19.7 & 89.6 G & 30.8 ms & 36.6/19.5 &49.7 G & 24.0 ms & 65.6 M \\
& SODFormer & TPAMI'22 & --- & --- & --- & --- & 50.4/20.7 & 62.5 G & 39.7 ms & 82.0 M \\
% & RENet & ICRA'23 & - & 37.3/22.2 & - & -& 54.9/28.8 & - & 14.2ms & 59.8M \\

 & EOLO & ICRA'24 & 65.1/37.8 & 33.9/19.6 & 13.7 G & 330.2 ms & 47.2/22.0 & 8.9 G & 326.4 ms & 21.5 M \\
& SFNet & ITS'24 & 80.0/50.9 & 51.4/30.4 & 209 G & 44.8 ms & 59.6/31.9 & 135.9 G & 42.3 ms & 57.5 M \\
% \hline
% \hdashline

& \makecell{Our pipeline \\+ CFM} & CVPR'24 & 86.4/53.1 & \textcolor{blue}{60.0}/35.3 & 256.0 G & 58.9 ms & 60.0/30.8 & 80.93 G & 17.2 ms &46.0 M \\
& \makecell{Our pipeline \\+ CAFR} & ECCV'24 & \textcolor{blue}{87.4}/54.8 & 59.9/34.6 & 208.5 G & 50.4 ms & \textcolor{blue}{61.2/31.8} &66.51 G & 18.9 ms &43.6 M \\
& \makecell{MCFNet (this study)} & --- & \textcolor{red}{90.7/60.9} & \textcolor{red}{67.4/41.8} & 227.8 G & 47.3 ms & \textcolor{red}{61.8/32.6} &72.68 G &18.8 ms &52.1 M \\
\Xhline{1.5pt}
\end{tabular}
}
\end{table*}

% our fusion strategy significantly outperforms existing methods, improving mAP50 by 3.3\% on the class-balanced and 7.4\% on the class-imbalanced splits of the DSEC-Det dataset. It proves that our CMM can enable the model to more accurately see the complementary relationship between modalities and achieve adaptive fusion. 
In addition, on the PKU-DAVIS-SOD dataset, our method outperforms the SODFormer \citep{li2023sodformer} by 11.9\% and 11.4\% in mAP and mAP50, respectively. As shown in Table \ref{tab:sod_comparison}, our model shows consistent performance advantages under various challenging conditions, outperforming SODFormer \citep{li2023sodformer} by 10.2\%, 9.6\%, and 11.8\% in mAP50 for normal, low-light, and motion blur subsets, respectively. These results indicate that our method achieves more robust object detection performance by accurate spatiotemporal alignment and adaptive fusion across modalities, especially in challenging traffic scenes with complex lighting conditions and motion blur.

\begin{table*}[h]
\belowrulesep=0pt
\aboverulesep=0pt
\centering
\caption{Performance comparison with SOTA method on different scene subsets of PKU-DAVIS-SOD dataset. The best and the second-best performances are marked in red bold and blue bold.}
\label{tab:sod_comparison}
\renewcommand\arraystretch{1.3}
\scalebox{0.95}{
% \begin{tabular}{cc|cc|cc|cc}
\begin{tabular}{cccccccc}
\Xhline{1.5pt}
\multirow{3}{*}{Method} & \multirow{3}{*}{Pub\&Year} & \multicolumn{6}{c}{PKU-DAVIS-SOD} \\
\cline{3-8}
& & \multicolumn{2}{c}{Normal} & \multicolumn{2}{c}{Motion\_blur} & \multicolumn{2}{c}{Low\_light} \\
\cmidrule(lr){3-4} \cmidrule(lr){5-6} \cmidrule(lr){7-8}
& & mAP & mAP50 & mAP & mAP50 & mAP & mAP50 \\
\hline
SODFormer & TPAMI'22 & 24.1 & 56.9 & 18.3 & 43.2 & 12.2 & 37.4 \\
SFNet & ITS'24 & 32.3 & 62.4 & 23.1 & 46.7 & 17.6 & 41.2 \\

Our pipeline + CFM & CVPR'24 & {33.6} & {65.1} & {25.0} & {50.9} & {21.2} & {48.6} \\
Our pipeline + CAFR & ECCV’24 & \color{blue}{34.4} & \color{blue}{66.6} & \color{blue}{26.0} & \color{blue}{51.7} & \color{red}{21.6} & \color{blue}{49.9} \\
MCFNet (this study) & --- & \color{red}{34.8}\textcolor{red}{+0.4} & \color{red}{67.1}\textcolor{red}{+0.5} & \color{red}{26.0} & \color{red}{52.8}\textcolor{red}{+1.1} & \color{red}{21.8}\textcolor{red}{+0.2} & \color{red}{49.2}\textcolor{blue}{--0.7} \\
\Xhline{1.5pt}
\end{tabular}
}
\end{table*}

\textbf{Cross-scene evaluation on PKU-DAVIS-SOD.}
To further demonstrate the robustness of our proposed method, we conduct a cross-scene validation. Specifically, we transferred a model trained solely on the DSEC-Det dataset to the PKU-DAVIS-SOD dataset. Due to the difference in the number of object categories between the two datasets, we applied only light fine-tuning to the detection head. As shown in Table~\ref{tab:migration}, despite differences in data distribution and class definitions between datasets, our model still outperformed other methods, demonstrating strong generalization and robustness across domains.
\begin{table*}[h]
\belowrulesep=0pt
\aboverulesep=0pt
\centering
\caption{Cross-scene performance comparison with SOTA methods on the PKU-DAVIS-SOD dataset. The best performance is marked with red bold.}
\label{tab:migration}
\renewcommand\arraystretch{1.3}
\scalebox{0.95}{
% \begin{tabular}{cc|cc}
\begin{tabular}{cccc}
\Xhline{1.5pt}
\multirow{2}{*}{Method} & \multirow{2}{*}{Pub\&Year} & \multicolumn{2}{c}{PKU-DAVIS-SOD} \\
\cline{3-4}
& & mAP & mAP50 \\
\hline
SFNet & ITS'24 & 19.7 & 41.9 \\
Our pipeline + CFM & CVPR'24 & 18.5 & 39.5 \\
MCFNet(this study) & --- & \text{\color{red}{24.7}} & \text{\color{red}{49.7}}\\
\Xhline{1.5pt}
\end{tabular}
}
\end{table*}

\textbf{Comparison with SOTA event frame accumulation methods and motion compensation methods.} To further demonstrate the superiority of our ECM representation method, we compare it with four Event frame accumulation methods: Timestamp \citep{park2016performance}, DiST \citep{kim2021n}, Voxel \citep{zihao2018unsupervised}, and TAF \citep{liu2023motion}.  Additionally, we evaluate five motion compensation methods: (i) model-based (MB): CMax \citep{gallego2018unifying}, ST-PPP \citep{gu2021spatio} and MCM \citep{shiba2022secrets}, (ii) self-supervised learning (SSL): ConGru-EV-FlowNet \citep{Paredes-Vallés_Hagenaars_Croon_2021, paredes2023taming} configuration with the best performance. For the motion compensation method, the resulting IWE is used as input for our detection network. We keep the detection network, upsampling module, and fusion module the same and only change the input of event modality. The results are reported in Table \ref{tab:event}.

\begin{table*}[h]
\belowrulesep=0pt
\aboverulesep=0pt
\centering
\caption{Performance comparison with state-of-the-art (SOTA) event representation methods and motion compensation methods. The best and second-best performances are marked in red bold and blue bold, respectively. MB: Model-based; SSL: Self-supervised learning.}
\label{tab:event}
\renewcommand\arraystretch{1.8}
\scalebox{0.65}{
% \begin{tabular}{cccc|cc|cc|cc}
\begin{tabular}{cccccccccc}
\Xhline{1.5pt}
& \multirow{2}{*}{{\Large Method}} & \multirow{2}{*}{{\Large Pub. \& Year}} & {{\Large Inference}} & \multicolumn{2}{c}{{\large DSEC-Det (Class-balanced)}} & \multicolumn{2}{c}{{\large DSEC-Det (Class-imbalanced)}} & \multicolumn{2}{c}{{\Large PKU-DAVIS-SOD}} \\
\cmidrule(lr){5-6} \cmidrule(lr){7-8} \cmidrule(lr){8-9}
& & & {\Large time} & {\Large mAP} & {\Large mAP50} & {\Large mAP} & {\Large mAP50} & {\Large mAP} & {\Large mAP50} \\
\hline
\multirow{3}{*}{{\Large \rotatebox{90}{MB}}} 
& {\Large CMax} & {\Large CVPR'18} & {\Large 2.42s} & {\Large 60.0} & {\Large 89.6} & {\Large \color{blue}{\text{41.5}}} & {\Large 65.8} & {\Large 31.1} & {\Large 60.3} \\
& {\Large ST-PPP} & {\Large ICCV'21} & {\Large 2.48 s} & {\Large \color{blue}{\text{60.3}}} & {\Large 89.9} & {\Large 41.4} & {\Large \color{blue}{\text{67.0}}} & {\Large 31.2} & {\Large 60.2} \\
& {\Large MCM} & {\Large ECCV'22} & {\Large 65 s} & {\Large 60.1} & {\Large \color{blue}{\text{90.0}}} & {\Large 41.3} & {\Large 66.0} & {\Large 31.5} & {\Large 60.0} \\
\hdashline

\multirow{2}{*}{{\Large \rotatebox{90}{SSL}}} 
& \thead{\Large{ConvGRU-} \\ \Large{EV-FlowNet}} & {\Large NeurIPS'21} & {\Large 0.48 s} & {\Large 60.1} & {\Large 89.9} & {\Large 41.3} & {\Large 66.0} & {\Large \color{blue}{\text{32.4}}} & {\Large 62.0} \\
& {\Large Federico et al.} & {\Large ICCV'23} & {\Large 0.29 s} & {\Large 60.2} & {\Large 89.9} & {\Large 40.1} & {\Large 63.8} & {\Large \color{blue}{\text{32.4}}} & {\Large \color{red}{\text{62.3}}} \\
\hdashline

& {\Large TAF} & \thead{{\Large IEEE Trans.} \\ {\Large TIM'23}} & {\Large ---} & {\Large 60.2} & {\Large 89.6} & {\Large 39.7} & {\Large 63.3} & {\Large 31.0} & {\Large 59.9} \\
& {\Large Timestamp} & {\Large ICIP'16} & {\Large ---} & {\Large 59.7} & {\Large 89.3} & {\Large 40.5} & {\Large 65.6} & {\Large 31.7} & {\Large 60.9} \\
& {\Large Voxel} & {\Large CVPR'19} & {\Large ---} & {\Large 59.4} & {\Large 89.0} & {\Large 41.1} & {\Large 65.7} & {\Large 30.8} & {\Large 59.6} \\
& {\Large DiST} & {\Large ICCV'21} & {\Large ---} & {\Large 60.0} & {\Large 89.9} & {\Large \color{blue}{\text{41.5}}} & {\Large 65.6} & {\Large 31.4} & {\Large 60.7} \\
& {\Large ECM(this study)} & {\Large ---} & {\Large 0.03 s} & {\Large \color{red}{\text{60.9}}} & {\Large \color{red}{\text{90.7}}} & {\Large \color{red}{\text{41.8}}} & {\Large \color{red}{\text{67.4}}} & {\Large \color{red}{\text{32.6}}} & {\Large \color{blue}{\text{61.8}}} \\
\Xhline{1.5pt}
\end{tabular}
}
\end{table*}

Among all event frame accumulation methods, our ECM achieves optimal performance by predicting motion vectors to align timestamps, resulting in accurate temporal alignment and high-quality frames with sharp edges and reduced noise. For example, on the class-imbalanced DSEC-Det dataset, our event representation outperforms voxel \citep{zihao2018unsupervised} by 1.7\% on mAP50.

\begin{figure*}[h]
\makebox[\textwidth][c]{
    \includegraphics[width=7in]{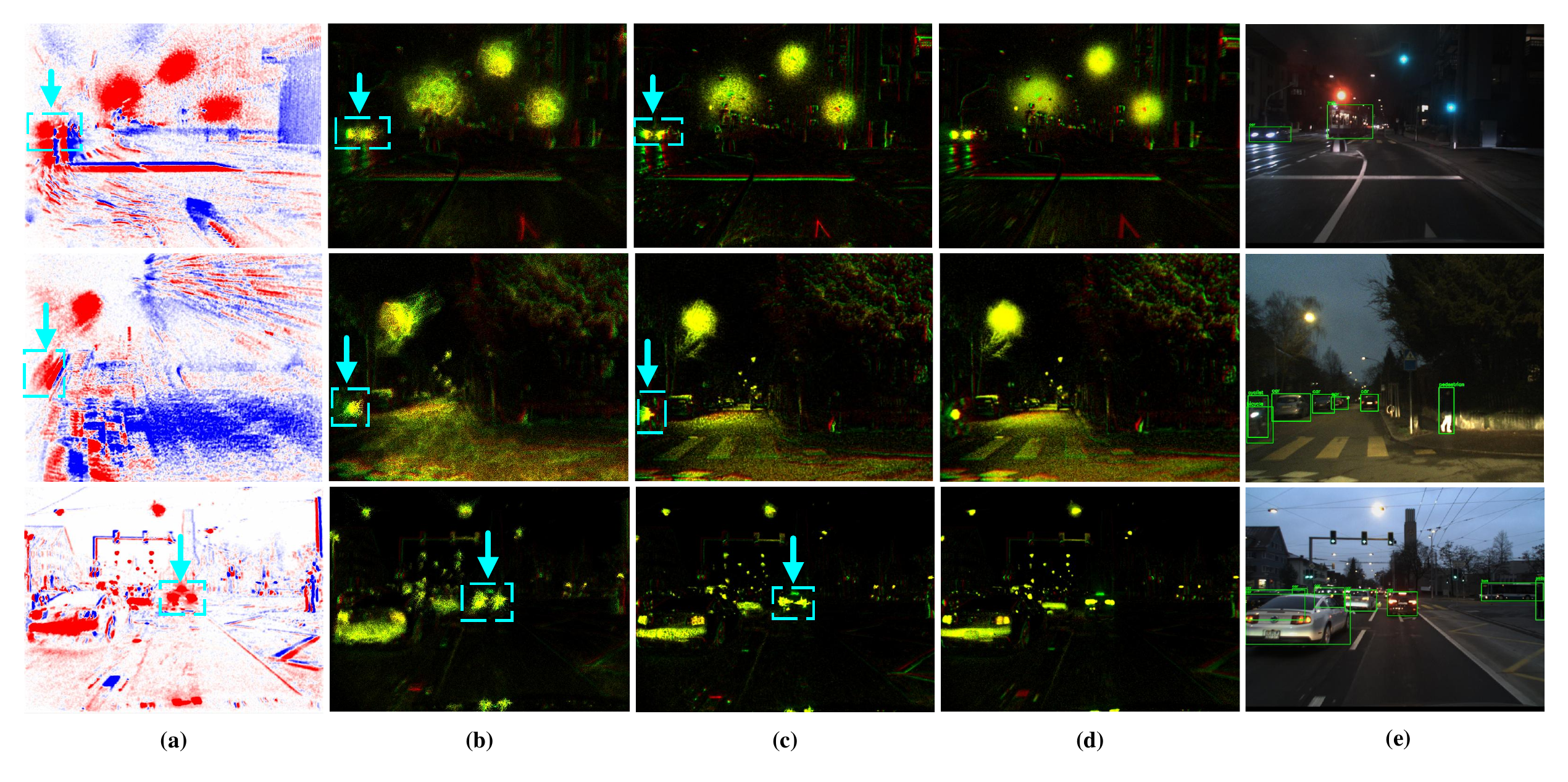}
}
\caption{Qualitative comparison with SOTA motion compensation event representation methods on the DSEC-Det dataset: (a) ST-PPP, (b) ConvGRU-Ev-FlowNet, (c) Ev-FlowNet, (d) ECM(this study), and (e) GT.}
\label{fig_4}
\end{figure*}

Compared to model-based and self-supervised motion compensation methods, our ECM also demonstrates superior performance. For instance, on the class-imbalanced DSEC-Det dataset, our event representation surpasses ConvGRU-EV-FlowNet \citep{Paredes-Vallés_Hagenaars_Croon_2021} by 1.4\% on mAP50. This proves that when jointly trained with detection networks in an end-to-end architecture, our ECM learns scene features that overcome the limitations of brightness constancy and linear motion assumptions, generating event representations that are better suited for various complex scenarios and beneficial for object detection tasks. In contrast, different motion states and frequent illumination changes adversely affect optical flow estimation, leading to artifacts and noise generation in IWE, which impacts detection performance, as shown in Fig. \ref{fig_4}.

\subsection{Qualitative results}
\begin{figure*}[h]
\makebox[\textwidth][c]{
    \includegraphics[width=6.5in]{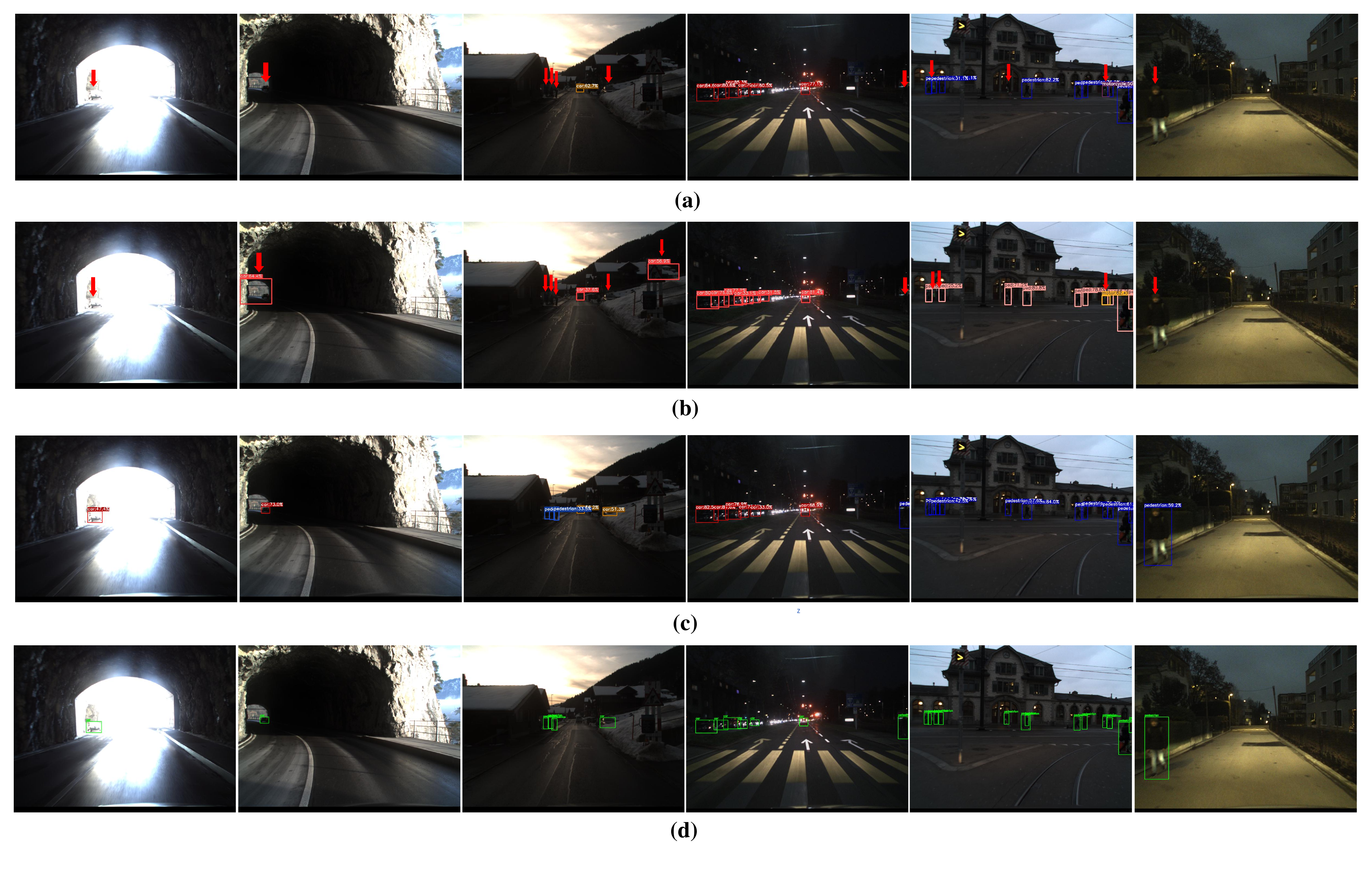}
}
\caption{Qualitative comparison with two leading SOTA detection methods on the class-imbalanced DSEC-Det dataset: (a) EvImHnadNet; (b) SFNet; (c) MCFNet(this study); and (d) GT.  We utilize {\color{red}red} arrows to mark the failed cases.}
\label{fig_3}
\end{figure*}

\textbf{Comparison with SOTA detection methods.}  We compare MCFNet with two leading SOTA methods Our pipeline+CFM \citep{jiang2024complementing} and SFNet \citep{liu2024enhancing}.  As shown in Fig. \ref{fig_3}, the first three columns show scenes with non-uniform exposure, while columns 4 and 5 show low-light multi-object scenes, and the last column is a motion blur scene. Our method achieves superior robustness compared to existing SOTA approaches in complex illumination conditions through accurate spatiotemporal alignment across modalities and excellent adaptive complementary fusion by simultaneously perceiving global information from both modalities in the scene, thereby enabling successful detection.

\textbf{Comparison with SOTA event representation methods.} We compare our event representation generated from ECM with two leading SOTA motion compensation approaches ST-PPP, ConvGRU-EV-FlowNet \citep{gu2021spatio,Paredes-Vallés_Hagenaars_Croon_2021} and representations generated by EvFlow without end-to-end joint training, as illustrated in Fig. \ref{fig_4}. The first row shows a scenario of rapidly moving vehicles with bright self-emitting lights in nighttime conditions.  The second row presents a scene where the ego vehicle's headlights illuminate the target vehicle while it simultaneously emits its own lighting during motion.  In the third scenario, an examination of consecutive frames reveals that the center vehicle's brake lights have just activated in the current frame.  These three scenarios collectively showcase environments with either rapid illumination variations or complex motion patterns. As illustrated in columns 2--4 of Fig. \ref{fig_4}, existing motion compensation methods often fail under such conditions as environmental changes violate their underlying assumptions of brightness constancy and linear motion, resulting in artifacts, noise, and target deformation. In contrast, our ECM-generated representations maintain target fidelity while leveraging event motion characteristics, guided by the optimization from the object detection task.

\subsection{Ablation study}
In this section, we present our ablation study results on the DSEC-Det dataset to validate the effectiveness of each component in our proposed MCFNet.

\textbf{Contribution of our MCFNet components.} We take dual-stream YOLOX architecture with Voxel Event representation and simple feature addition fusion as the baseline to perform a comprehensive ablation study and analyze the performance of each component in the MCFNet. Table \ref{tab:ablation} shows the results from different combinations of modules in our method. It can be observed that each proposed module contributes positively to the detection performance of MCFNet. Especially on the class-imbalanced DSEC-Det dataset, our model achieves significant improvements over the baseline, with increases of 8.1\% and 12\% in mAP and mAP50 metrics respectively, demonstrating the effectiveness of each module in dynamic traffic scenarios.

\begin{table}[h]
\belowrulesep=0pt
\aboverulesep=0pt
\centering
\caption{Ablation study of proposed components. \textcolor{red}{Red bold} indicates improvement from baseline.}

\label{tab:ablation}
\renewcommand\arraystretch{1.5}
\scalebox{0.85}{
% \begin{tabular}{ccc|cc|cc}
\begin{tabular}{ccccccc}
% \hline
\Xhline{2pt}
{\multirow{2}{*}{\large{ECM}}} &{\multirow{2}{*}{\large{EDUM}}} & {\multirow{2}{*}{\large{CMM}}} & \multicolumn{2}{c}{\large{DSEC-Det class-balanced}} & \multicolumn{2}{c}{\large{DSEC-Det class-imbalanced}} \\
\cmidrule(lr){4-5} \cmidrule(lr){6-7} 
&&& \large{mAP} & \large{mAP50} & \large{mAP} & \large{mAP50} \\
\hline
\multicolumn{3}{c}{\large{YOLOX(voxel)}} & \large{53.7} & \large{82.5} & \large{33.7} & \large{55.4} \\
\hline
\large{\checkmark} &{---} & {---}& \large{53.8}{\color{red}+0.1} & \large{82.7}{\color{red}+0.2} & \large{34.1}{\color{red}+0.4} & \large{55.6}{\color{red}+0.2} \\
\large{\checkmark} &{---} & \large{\checkmark} & \large{55.9}{\color{red}+2.2} & \large{85.9}{\color{red}+3.4} & \large{34.0}{\color{red}+0.3} & \large{56.9}{\color{red}+1.5} \\
\large{\checkmark} & \large{\checkmark} &{---} & \large{60.0}{\color{red}+6.3} & \large{89.4}{\color{red}+6.9} & \large{40.4}{\color{red}\small{+6.7}} & \large{64.7}{\color{red}+9.3} \\
\large{\checkmark} & \large{\checkmark} & \large{\checkmark} & \large{\text{60.9}}{\text{\color{red}+7.2}} & \large{\text{90.7}}{\text{\color{red}+8.2}} & \large{\text{41.8}}{\text{\color{red}+8.1}} & \large{\text{67.4}}{\text{\color{red}+12.0}} \\
% \hline
\Xhline{1.5pt}
\end{tabular}
  }

\end{table}

\begin{figure*}[h]
\makebox[\textwidth][c]{
    \includegraphics[width=6in]{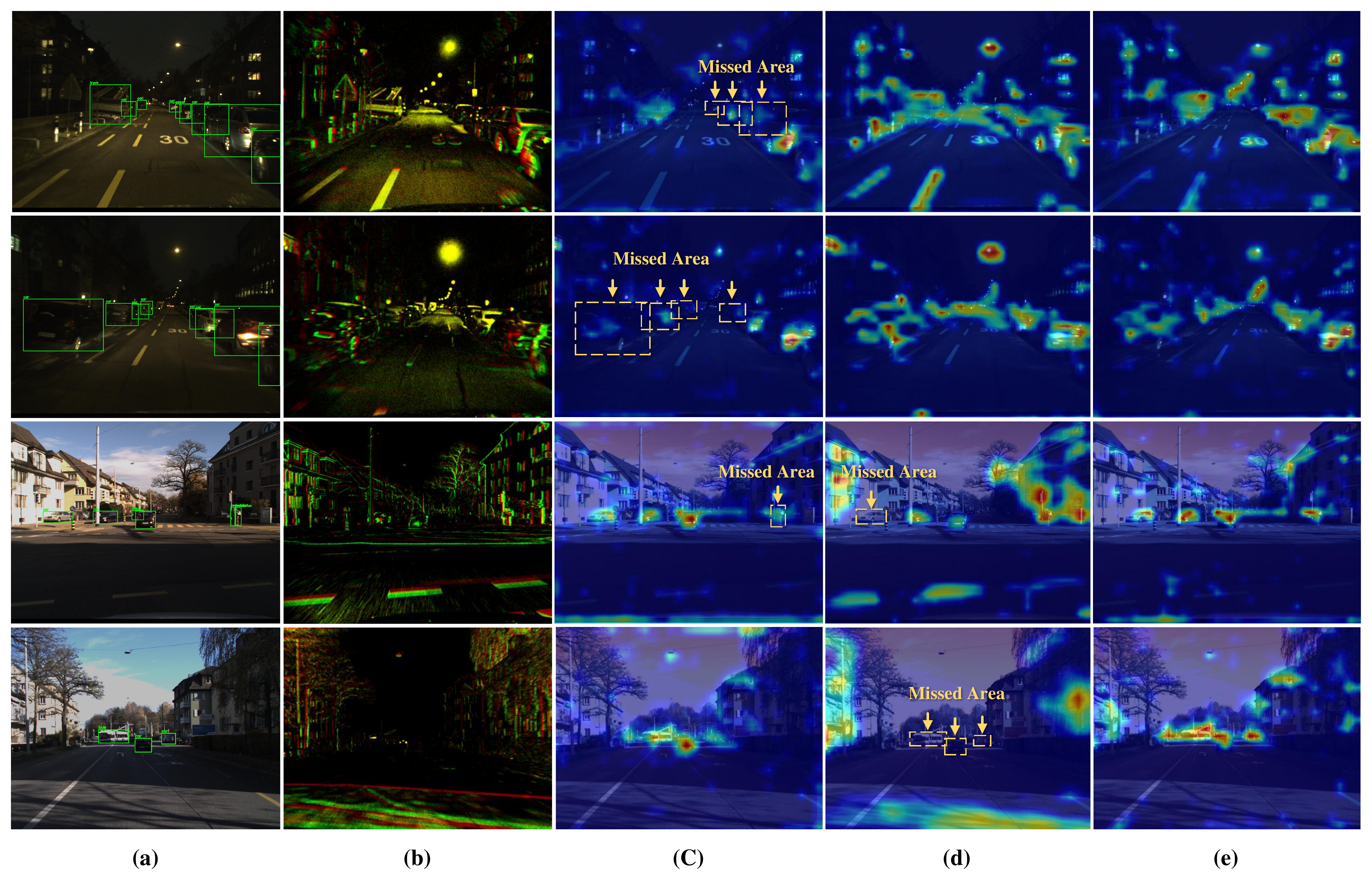}
}
\caption{Visualization of features before and after CMM. (a) GT representing the scene, (b) Event representation corresponding to the scene, (c) feature visualization of RGB modality before it passes through CMM, (d) feature visualization of event modality before it passes through CMM, and (e) fused feature visualization after CMM module. The {\color{red}} boxes indicate the unfocused target areas.}
\label{fig_5}
\end{figure*}

The feature maps before and after applying CMM are shown in Fig. \ref{fig_5}. As observed in the first and second rows depicting nighttime scenes in Fig. \ref{fig_5}, the event modality exhibits more stable imaging capabilities in low-light conditions compared to the RGB modality, due to its high dynamic range, enabling more effective target information capture. Conversely, in the daytime scenes shown in the third and fourth rows, when the ego vehicle moves simultaneously with target objects, events become sparse due to relative stasis or minimal relative motion. Meanwhile, RGB images under favorable lighting conditions contain rich chromatic, textural, and semantic information, facilitating more effective target capture. After CMM fusion, the model simultaneously models global information from both modalities to precisely identify inter-modal complementary relationships, selectively emphasizing key information through adaptive fusion to achieve complementary modal advantages and capture all target information in the scene.

\textbf{Benefits of joint training of ECM.}  We validate the effectiveness of our joint training strategy on DSEC-Det \citep{gehrig2021dsec}. The two-stage training strategy refers to a configuration where the optical flow network parameters remain fixed without participating in the backpropagation process during detector training.  Our ECM employs end-to-end joint network training, where event correction optimization is guided by detection results.  As shown in the Table \ref{tab:training_comparison}, our ECM learns scene features that surpass the assumptions of brightness constancy and linear motion, achieves more accurate temporal alignment, and generates event representations better aligned with object detection requirements, thereby delivering superior performance.
\begin{table}[h]
\belowrulesep=0pt
\aboverulesep=0pt
\centering
\caption{Comparing the results of EvFlowNet without joint training with ECM.}
\label{tab:training_comparison}
\renewcommand\arraystretch{1.5}
\scalebox{0.9}{
% \begin{tabular}{c|c|cc}
\begin{tabular}{cccc}
% \hline
\Xhline{1.5pt}
{\multirow{2}{*}{\large{Method}}} & {\multirow{2}{*}{\large{Training strategy}}}& \multicolumn{2}{c}{\large{DSEC-Det class-imbalanced}} \\
\cline{3-4}
& & \large{mAP} & \large{mAP50} \\
\hline
\large{EV-FlowNet} & \large{Two-stage} & \large{41.1} & \large{65.4} \\
\large{ECM (this study)} & \large{Joint training} & \text{\large{41.8}} & \text{\large{67.4}} \\
% \hline
\Xhline{1.5pt}
\end{tabular}
}
\end{table}

\textbf{Selection of upsampling methods.} To validate the effectiveness of our dynamic upsampling method EDUM, we conducted comparative experiments against common upsampling approaches including Pixel Shuffle and transposed convolution. As demonstrated in Table \ref{tab:upsample}, dynamically adjusting weights based on input enables the model to learn pixel distribution variations across different scenarios, thereby achieving more precise spatial alignment and enhanced detection performance.
\begin{table*}[h]
\belowrulesep=0pt
\aboverulesep=0pt
\centering
\caption{Performance comparison of different upsampling methods.}
\label{tab:upsample}
\renewcommand\arraystretch{1.5}
\scalebox{0.85}{
\begin{tabular}{ccc}
\Xhline{1.5pt}
\multirow{2}{*}{\large{Upsampling Method}} & \multicolumn{2}{c}{\large{DSEC-Det (class-imbalanced)}} \\
\cline{2-3}
& \large{mAP} & \large{mAP50} \\ 
\hline
\large{Pixel Shuffle} & \large{40.7} & \large{65.2} \\
\large{Transposed conv (3$\times$3)} & \large{41.5} & \large{65.8} \\
\large{EDUM (this study)} & \text{\large{41.8}} & \text{\large{67.4}} \\
\Xhline{1.5pt}
\end{tabular}
}
\end{table*}
% \IEEEpubidadjcol
% \section{Tables}
% Note that, for IEEE-style tables, the
%  $\backslash${\tt{caption}} command should come BEFORE the table. Table captions use title case. Articles (a, an, the), coordinating conjunctions (and, but, for, or, nor), and most short prepositions are lowercase unless they are the first or last word. Table text will default to $\backslash${\tt{footnotesize}} as
%  the IEEE normally uses this smaller font for tables.
%  The $\backslash${\tt{label}} must come after $\backslash${\tt{caption}} as always.
 
% \begin{table}[!t]
% \caption{An Example of a Table\label{tab:table1}}
% \centering
% \begin{tabular}{|c||c|}
% \hline
% One & Two\\
% \hline
% Three & Four\\
% \hline
% \end{tabular}
% \end{table}

% \section{Algorithms}
% Algorithms should be numbered and include a short title. They are set off from the text with rules above and below the title and after the last line.

\section{Conclusion}
In this study, we propose a Motion Cue Fusion Network (MCFNet) for robust object detection in dynamic traffic scenarios. ECM overcomes the limitations of the traditional assumptions of constant illumination and linear motion in the optical flow estimation task by being effectively guided by the object detection task to obtain high-quality event frames. EDUM dynamically upsampling event features based on the feature space distribution, aligning the resolutions of the two modalities while maximizing the use of high-quality information. CMM enhances the accuracy of the model's perception and selection of dominant modal features in different scenarios by simultaneously performing inter-modal feature interaction and global information extraction, thereby achieving adaptive fusion. Experiments show that our MCFNet significantly outperforms existing methods in various complex and dynamic traffic scenarios. 

Although the proposed method achieves excellent performance in detection accuracy, the introduction of spatiotemporal alignment for event data and the multimodal cross-fusion mechanism results in a relatively complex model structure with high computational overhead, posing challenges for deployment on resource-constrained devices. In future work, we plan to introduce sparsity constraints on event data. For example, using sparse token mechanisms to filter out inactive regions and perform fusion only where events occur, thereby reducing redundant computation. We also aim to explore dynamic token selection and model pruning to reduce computational load further and accelerate inference for real-time applications.

% 正确的参考文献调用方式
% \bibliography{07-reference} 
\section*{References}
\bibliographystyle{elsarticle-harv}
\bibliography{main}

\section*{Author biography}
\renewcommand{\baselinestretch}{1}  % 修改行间距，增加文字间距
\begin{biography}[figures/Zhanwen-Liu]{Zhanwen Liu}received the B.S. degree from Northwestern Polytechnical University, Xi'an, China, in 2006, the M.S. and the Ph.D. degrees in traffic information engineering and control from Chang'an University, Xi'an, China, in 2009 and 2014, respectively. She is currently a Professor with School of Information Engineering, Chang’an University. Her research interests include motion perception, behavior prediction, and data-closed-loop autonomous driving testing.
\end{biography}

\begin{biography}[YujingSun]{Yujing Sun}received the B.S. degree from Chang'an University, Xi'an, China in 2023 and is currently pursuing the M.S. degree in computer science and technology in Chang'an University. Her current research interests include object detection and optical flow estimation based on event cameras and their applications in autonomous driving perception.
\end{biography}

\begin{biography}[figures/Yang-Wang]{Yang Wang} received the B.S. degree from Chang'an University, Xi'an, China, in 2016, and the Ph.D. degree in control science and engineering from the University of Science and Technology of China, in 2021. He is currently an Associate Professor with the School of Information Engineering, Chang'an University, Xi'an, China. His central research interests focus on computer vision and multimedia processing.
\end{biography}

\begin{biography}[figures/Nan-Yang]{Nan Yang}received the B.S. degree from Chang'an University in Xi'an, China, in 2022, where he is currently working toward the Ph.D. degree in traffic information engineering and control. His current research interests include object detection and multiple-object tracking based on event cameras, and their applications in intelligent vehicle and road infrastructure perception.
\end{biography}

\begin{biography}[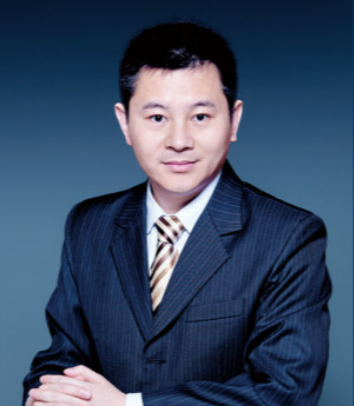]{Shengbo Eben Li}received the M.S. and Ph.D. degrees from Tsinghua University in 2006 and 2009, respectively. Before joining Tsinghua University, he has worked at Stanford University, University of Michigan, and UC Berkeley. His active research interests include intelligent vehicles and driver assistance, deep reinforcement learning, optimal control and estimation, etc.
\end{biography}

\begin{biography}[figures/Xiangmo-Zhao]{Xiangmo Zhao}received the Ph.D. degree from Chang’an University, Xi’an, China, in 2003. He is currently a Distinguished Professor with the School of Information Engineering, Chang’an University, and also is the President of Xi'an University of Architecture and Technology. He currently serves as the Vice Chairman of the China Society of Automotive Engineers. His current research interests include testing of connected vehicles, automated vehicles, and intelligent transportation systems. 
\end{biography}

% \begin{biography}[{\includegraphics[width=1in,height=1.25in,clip,keepaspectratio]{figures/athor5.png}}]{Xiangmo Zhao}

\end{document}